\documentclass[a4paper]{article}

\usepackage{INTERSPEECH2018}
\newcommand{\ee}{end-to-end }

\title{End-to-end named entity extraction from speech} 
\name{Sahar Ghannay$^1$, Antoine Caubri\`{e}re$^1$, Yannick Est\`{e}ve$^1$, Antoine Laurent$^1$, Emmanuel Morin$^2$}
\address{
  $^1$LIUM - University of Le Mans, France\\
  $^2$LS2N - University of Nantes, France}
\email{$^1$firstname.lastname@univ-lemans.fr, emmanuel.morin@univ-nantes.fr}

\begin{document}

\maketitle
\begin{abstract}
Named entity recognition (NER) is among SLU tasks that usually extract  semantic information from textual documents. 
Until now, NER from speech is made through a pipeline process that consists in processing first an automatic speech recognition (ASR) on the audio and then processing a NER on the ASR outputs.
Such approach has some disadvantages (error propagation, metric to tune ASR systems sub-optimal in regards to the final task, reduced space search at the ASR output level,...) and it is known that more integrated approaches outperform sequential ones, when they can be applied.
In this paper, we present a first study of \ee approach that directly extracts named entities from speech, though a unique neural architecture. On a such way, a joint optimization is able for both ASR and NER. Experiments are carried on French data easily accessible, composed of data distributed in several evaluation campaign.
Experimental results show that this end-to-end approach provides better results (F-measure=0.69 on test data) than a classical pipeline approach to detect named entity categories (F-measure=0.65).

\end{abstract}

\section{Introduction}

Named entities are sequences of words that bring basic predefined semantic information that usually refers to locations, persons, organization\ldots that can be denoted by proper nouns or that are unique in the real world, and they usually include numeric and temporal values. Named entities often constitute the first semantic bricks to extract in order to construct a structured semantic representation of a document content. 

Named entity recognition (NER) is among SLU tasks that usually extract semantic information from textual documents. 
Until now, NER from speech is made through a pipeline process that consists in processing first an automatic speech recognition (ASR) on the audio and then processing a NER on the ASR outputs.
Such approach has some disadvantages.

For instance, ASR errors have a negative impact on the NER performances, introducing noise within the text to be processed~\cite{hatmi2013named}. Rule-based NER systems are usually built to process written language and are not robust to ASR errors. Machine learning based systems do not have good performance when they are trained on perfect transcriptions and deployed to process ASR ones, even if that can be partially compensated by simulating ASR errors in textual training data~\cite{simonnet:hal-01715923}.
Additionally, ASR systems are generally tuned in order to get the lowest word error rate on a validation corpus, but this metric is not optimal to the NER task. For instance, this metric does not distinguish between errors on verbs or proper nouns while such errors do not have the same impact for NER. To compensate this problem, some dedicated metrics to tune ASR systems for better NER performances have been proposed, such as in~\cite{jannet2017investigating}. Another inconvenience is that usually no information about named entities are used in the ASR process, while such information could help to better choose the partial recognition hypotheses that are dropped away during the decoding process. As a consequence, even when confusion networks or word lattices are used to go beyond the 1-best ASR hypothesis for a better robustness to ASR errors~\cite{hakkani2006beyond}, such search space have been pruned without taking into account knowledge on named entity.

In the past, and integrated approach built on a high coupling of ASR and NER modules has been proposed~\cite{servan2006conceptual}, based on the finite-state machine (FSM) paradigm (\textsl{i.e.} transducer composition), showing that such integration can offer significant improvements in terms of NER quality. The main limit of this approach concerns the FSM paradigm itself, that is not able to natively model long distant constraint without combinatory explosion and that, by nature, can only express dependencies through a regular grammar. Another proposition to inject information about named entities in the ASR consists in directly adding some expressions of named entities into the ASR vocabulary~\cite{hatmi2013incorporating}, and to estimate a language model for speech recognition that take into account these named entity expressions. The main default of a such approach is that it cannot allow to detect named entity that were not injected in the ASR vocabulary.

All of these issues motivate our research work on neural \ee approach to extract named entities from speech.  
On a such way, a joint optimization is able for both ASR and NER in a NER task perspective, the architecture is more compact than the ones used in usual pipeline, and we expect to take benefit of the deep neural architecture capacities to capture long distant constraint at the sentence level. 
Very recently, a similar approach has been proposed by Facebook on a paper posted on the arXiv.org website~\cite{serdyuk2018towards}. This \ee approach is dedicated to domain and intent classification tasks, and experiments were carried on internal data close to the spirit of the ATIS corpus, as expressed by the authors.

In this paper, we present a first study of an \ee approach to extract named entities.
Our neural architecture is very similar to the Deep Speech 2 neural ASR system proposed by Baidu in~\cite{amodei2016deep}. To use it for named entity recognition, we apply a multi-task training and modify the sequence of characters to be recognized from speech.
Experiments were carried on French data easily accessible, and so reproducible, that were distributed in the framework on evaluation campaigns and are still available. 
This paper is structured as follows. Section 2 describes the neural ASR architecture we used. Section 3 explains how we propose to exploit a such neural architecture for named entity extraction from speech. Section 4 presents some propositions to optimize the system and also compensate the lack of manually annotated audio data. Section 5 presents our experimental results, before the conclusion.


\section{Model architecture}

\begin{figure}[!htbp]
\begin{center}
\includegraphics [scale=0.45]{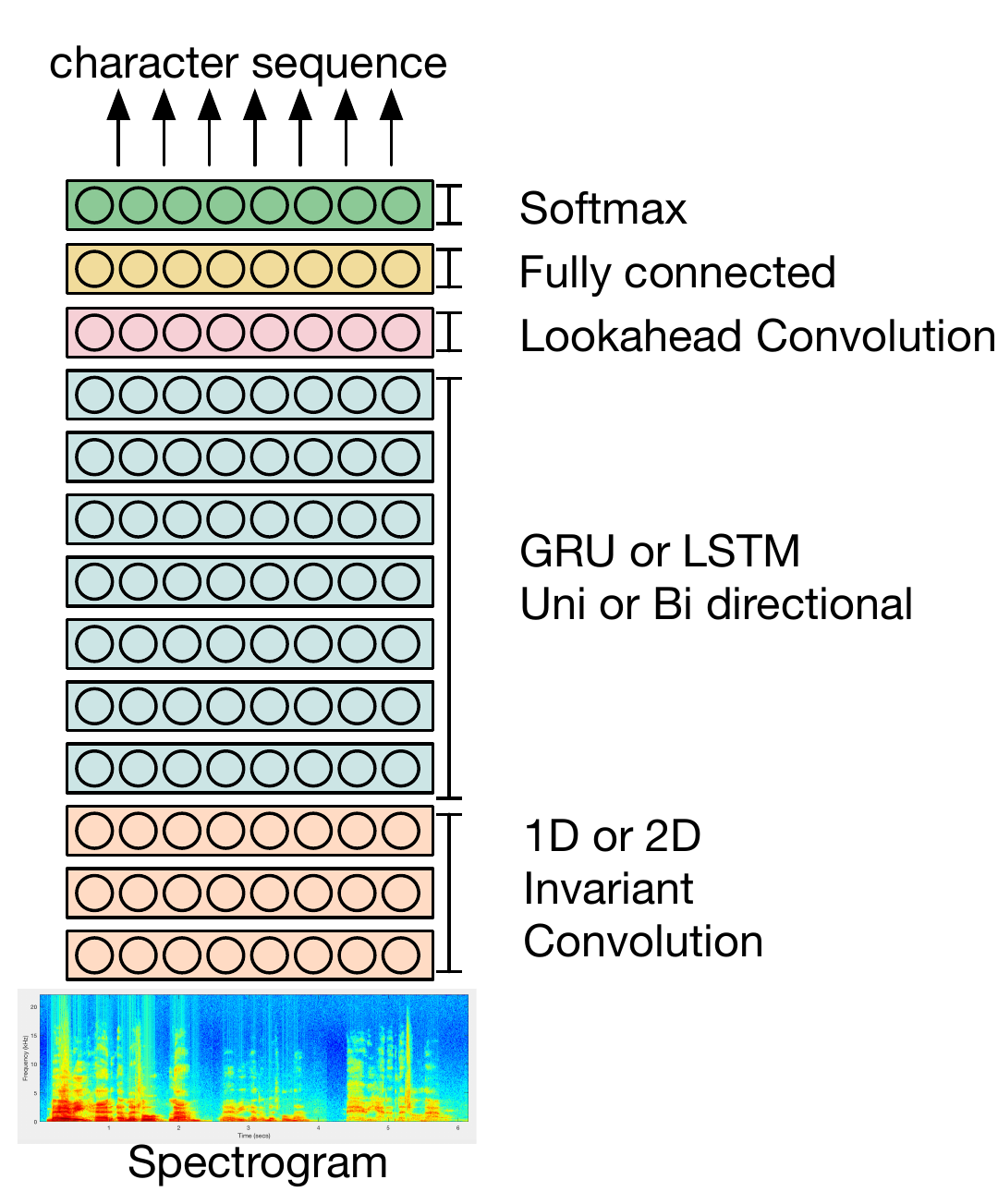}
\caption{\label{DeepSpeech} Deep RNN architecture used to extract named entities from French speech.}
\end{center}
\end{figure}

The RNN architecture used in this study is similar to the Deep~Speech~2 neural ASR system proposed by Baidu in~\cite{amodei2016deep}. 
This architecture is composed of $nc$ convolution layers (CNN), followed by $nr$ uni or bidirectional recurrent layers, a lookahead convolution layer~\cite{wang2016lookahead}, and one fully connected layer just before the softmax layer, as shown in Figure~\ref{DeepSpeech}.
\vspace{-0.2cm}
\vspace{-0.3cm}
The system is trained \ee using the CTC loss function~\cite{graves2006connectionist}, in order to predict a sequence of characters from the input audio. 
In our experiments we used two CNN layers and six bidirectional recurrent layers with batch normalization as mentioned in~\cite{amodei2016deep}. 

Given an utterance $x^{i}$ and label $y^{i}$  sampled from a training set $X ={(x^{1}, y^{1}), (x^{2}, y^{2}), . . .}$, the RNN architecture has to train to convert an input sequence  $x^{i}$ into a final transcription $y^{i}$s. For notational convenience, we drop the superscripts and use $x$ to denote a chosen utterance and $y$ the corresponding label.

The RNN takes as input an utterance $x$ represented by a sequence of log-spectrograms of power normalized audio clips, calculated on 20ms windows. As output, all the characters $l$ of a language alphabet may be emitted, in addition to the space character used to segment character sequences into word sequences (space denotes word boundaries).

The RNN makes a prediction $p(l_{t}|x)$ at each output time step $t$.  

At test time, the CTC model is coupled with a language model trained on a big textual corpus. A specialized beam search CTC decoder~\cite{hannun2014first} is used to find the transcription $y$ that maximizes :
\begin{equation}
Q(y)=log(p(l_{t}|x)) + \alpha log(pLM(y)) + \beta wc(y)
\end{equation}
where wc(y) is the number of words in the transcription $y$. The weight $\alpha$ controls the relative contributions of the language model and the CTC network. The weight $\beta$ controls the number of words in the transcription.

\section{Named entity extraction process} 
In the literature, many studies focus on named entity recognition from text.  
State-of-the-art systems are based on neural networks architectures. Some of them rely heavily on hand-crafted features and domain-specific knowledge~\cite{collobert2011natural,chiu2015named}. Recent approaches~\cite{lample2016neural,ma2016end} takes benefits from both word and/or character-level embeddings learned automatically, by using combination of bidirectional LSTM, CNN and CRF. 
However, named entities recognition from automatic transcriptions is less studied. 
This task is made through a pipeline process that consists in processing first an automatic speech recognition (ASR) on the audio and then processing a NER on the ASR outputs~\cite{raymond2013robust}. 
Usually, the named entity recognition task is to assign a named entity tag to every word in a sentence. A single named entity could concern several words within a sentence. For this reason, the word-level labels begin-inside-outside (BIO) encoding~\cite{Lance1995BIO} is very often adopted.

In this preliminary study, we focus on named entity extraction from speech using the network described above, without changing the neural architecture. 
We would like to evaluate if this neural architecture is able to capture high level semantic information that allow it to recognize named entities. For that, we propose to modify the character sequence that the neural network has to produce: information about named entities are added in the initial character sequence.
Instead of applying a BIO approach, we propose to add some tag characters in this sequence to delimit named entities boundaries, but also their category. 
We are interested to eight NE categories that are: \textit{person}, \textit{function}, \textit{organization}, \textit{location}, \textit{production}, \textit{amount}, \textit{time} and  \textit{event}. 

In our experiments, the system will attribute a tag ``$B_{NE}$'' or ``end''  only before and after the named entities, the other words are not concerned.  
To distinguish the named entity category, we consider a begin tag for each NE category. Only one ``end'' tag is used for all the NE categories, considering that since there is no overlap between named entities in a such representation, this information is sufficient to delimit the end of a named entity. 

\begin{figure*}[!htbp]
\begin{center}
\includegraphics [scale=0.45]{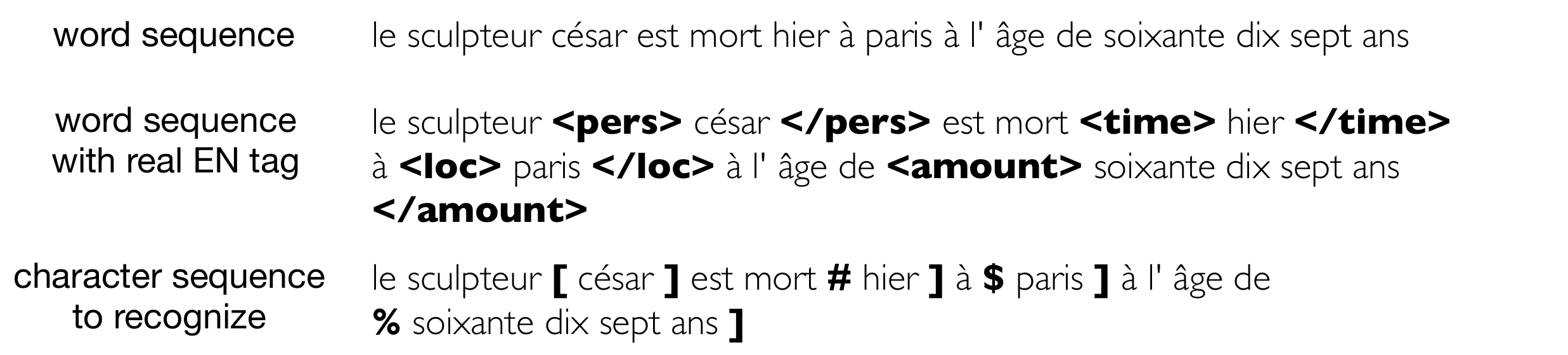}
\caption{\label{fig:example} Example of mapping the real NE tags to character sequence. This sentence means, in English and case sensitive: "the sculptor Caesar died yesterday in Paris at the age of seventy-seven years"}
\vspace{-0.1cm}
\end{center}
\end{figure*}

According to the eight named entity categories targeted by the task, nine NE tags has to be added to the character list emitted by the neural network: ``$B_{pers}$'', ``$B_{func}$'', ``$B_{org}$'', ``$B_{loc}$'', ``$B_{prod}$'', ``$B_{amount}$'', ``$B_{time}$'', ``$B_{event}$'', and $``end''$. 
As such neural model predicts a character at each time step, we propose to map each of these nine NE tags to one single special character, that corresponds respectively to: ``['', ``('', ``\{'',``\$'', ``\&'', ``\%'', ``\#'', ``)'' and ``]'', as illustrated in  Figure~\ref{fig:example}. 

With this way, the NE tags are included in the prediction process, and are taken into account by the CTC loss function during the training process.
\vspace{-0.2cm}
\section{Multi-task training, data augmentation, and starred mode} 
\label{sec:tricks}

Audio recordings with both manual transcriptions and manual annotations of named entities are relatively rare, while neural end-to-end approaches are known to need large amount of data to become competitive.

To compensate this lack of data, we first propose to apply a multi-task learning approach to train the neural network. This consists in starting to train it only for the ASR task, without emitting character used to represent named entities, on all the audio recordings available with their manual transcriptions. At the end, the softmax layer is reinitialized to take into consideration the named entity tag markers, and a new training process is realized, on the named entity recognition task, with only training data with manual annotations of named entities.

A second proposition consists in artificially increasing the training data for the named entity recognition task. For this purpose, we propose to apply a named entity recognition system dedicated to text data in order to tag the manual transcriptions used to train the ASR neural network. Then, these manual transcriptions automatically annotated with named entities can be injected in the training data used to train the neural network to extract named entities from speech.

In addition, since we want the system to focus on named entities, and since the CTC loss gives the same importance to each character, we propose to modify the character sequence that the neural network must emit to give more importance to named entities. This proposition is interesting to better understand how the CTC loss behaves on this case, and consists in replacing by a star ''*'' all character subsequences that do not contain a named entity. For instance, the character sequence presented in Figure~\ref{fig:example} becomes: \textit{* [ c\'esar ] * \# hier ] * \$ paris ] * \% soixante dix sept ans ]}. We call this approach the starred mode, and we expect that it can make the neural model more sensitive to named entities.
\vspace{-0.1cm}
\section{Experiments}
\subsection{Experimental setups}
\label{sec:corpus}
Experiments have been carried out on four different French corpora, including ESTER 1\&2, ETAPE and Quaero.
These corpora are composed of data recorded from francophone radio and TV stations, and are annotated with named entities.

The ESTER corpora were divided into three parts: training, development and evaluation.
ESTER 1 \cite{galliano2005ester} training (73 hours) and development (17 hours) corpora are composed of data recorded from four radio stations in French. ESTER 1 test corpus is composed of 10 hours coming from the same four radio stations plus two other stations, all of which recorded 15 month after the development data.

ESTER 2 \cite{galliano2009ester} training corpus was not annotated with named entities and was not used in this study. The development (17 hours) and test set (10 hours) is composed of manual transcriptions of speech recorded from six radio stations (two of those radio stations were already used in ESTER 1).

The ETAPE \cite{gravier2012etape} data consists of manual transcriptions and annotations of TV and radio shows. It contains 36 hours of speech, recorded between 2010 and 2011, divided into three parts: training (22 hours), development (7 hours) and test (7h).

QUAERO (ELRA-S0349) data is composed of 12 hours of manual transcriptions of TV and radio shows coming from 6 different sources recorded in 2010.

Our corpus, called DeepSUN, is the combination of those four corpora. 
The training corpus is composed of the training sets of ESTER 1, ETAPE and QUAERO, while, the  development and test sets are composed respectively of the development and test sets of ESTER 1\&2, and ETAPE.
It contains almost 160 hours of speech (training 107 hours, test 24 hours, development 30 hours). 
The distribution of named entities by categories in the corpus is summarized in Table \ref{tab:deepsun_guideline}. 
\vspace{-0.1cm}
\begin{table}[th]
 \caption{Distribution of named entities by categories in the DeepSUN corpus}
 \label{tab:deepsun_guideline}
 \centering
 \begin{tabular}{ cccc }
   \toprule
   \multicolumn{1}{c}{\textbf{}} & \multicolumn{3}{c}{\textbf{DeepSUN}} \\
   \multicolumn{1}{c}{\textbf{category}} & \multicolumn{1}{c}{\textbf{dev}} & \multicolumn{1}{c}{\textbf{test}} & \multicolumn{1}{c}{\textbf{train}}\\
   \midrule
   pers   & 6719   & 4766 	& 22115 	~~~ \\
   func   & 1830 	& 1425  & 6628 	~~~ \\
   org    & 5133 	& 3506 	& 15804 ~~~ \\
   loc    & 5195 	& 3915 	& 18159 ~~~ \\
   prod   & 652 	& 606 	& 2317 	~~~ \\
   time   & 3763	& 2769 	& 12020 	~~~ \\
   amount & 1591	& 1450 	& 5959 	~~~ \\
	event  & 79 	& 0 	& 321 	~~~\\
	 \midrule
    \multicolumn{1}{c}{\textbf{Sum}} &
   		\multicolumn{1}{c}{\textbf{24962}} & \multicolumn{1}{c}{\textbf{18437}} & \multicolumn{1}{c}{\textbf{83323}} \\
	\bottomrule
	\end{tabular}
\end{table}
\vspace{-0.1cm}

The performance of our approach is evaluated in terms of precision (P), recall(R) and F-measure for named entity detection, the named entity/value detection 
and the accuracy of the value detection when the named entities tags are correctly detected. 
These evaluations are made with the help of the \textit{sclite}\footnote{http://www.icsi.berkeley.edu/Speech/docs/sctk-1.2/sclite.htm} tool.
\subsection{Multi-task training}
For multi-task training, we first train the E2E architecture only for ASR task, without emitting character used to represent named entities. 
The system is trained on all the audio recordings available with their manual transcriptions around 297.7 hours of training set, including the data described above.  

It composed of two convolution layers and six BLSTM layers with batch normalization, the number of epochs was set to $35$.
This system achieves 20.70\% word error rate (WER) and
8.01\% character error rate (CER) on dev corpus (30.2 hours) and 19.95\% of WER and 7.68\% of CER on test set (40.8 hours). These results were obtained by applying a CTC beam search decoding coupled with a trigram language model. 
 Once this system is trained, the softmax layer is reinitialized to take into consideration the named entity tag markers, and a new training process is realized, on the named entity recognition task, with only training data with manual annotations of named entities described in table~\ref{tab:deepsun_guideline}.
\textbf{ In addition, for the training of both E2E and ASR systems, each training audio samples is randomly perturbed in gain and tempo for each iteration.}

\subsection{Experimental results}

We present in this section some experimental results. Table~\ref{tab:4G_etiquettes} shows the performances of the end-to-end model (E2E) to detect EN categories (among the eight ones). That means that in this evaluation we do not take care of values associated to the detected EN. The starred mode is also experimented and is called (E2E*) in the table: this mode provides better results in this task than the normal mode.

\begin{table}[!htbp]
\vspace{-0.1cm}
 \caption{Named entity category detection results for E2E and  E2E* (starred mode) systems} 
  \label{tab:4G_etiquettes}
  \centering
\begin{tabular}{ |l|c|c|c|c| }
   \hline
    \multicolumn{1}{|c|}{\textbf{System}} &\textbf{Corpus} & \multicolumn{1}{|c|}{\textbf{Precision}} & \multicolumn{1}{|c|}{\textbf{Recall}} & \multicolumn{1}{|c|}{\textbf{F-measure}}    \\
    \hline
    \hline
        E2E & dev   & \textbf{0.85}&0.57& 0.68  \\

    \hline
	E2E & test  & \textbf{0.83} & 0.52& 0.64  \\
    \hline
    \hline
     E2E* &dev   & 0.75 &  \bf{0.65}&  \bf{0.71}  \\
     \hline
     E2E* &test  &0.82 & \bf{0.57}&   \bf{0.67}  \\
    \hline
  \end{tabular}
\end{table} 

\vspace{-0.1cm}
Table~\ref{tab:4G_val_etiquettes} evaluates the quality of the category/value pairs that have been recognized. While precision and recall do not have the same behavior between normal and starred mode, both modes gets the same F-measure value.

\begin{table}[!htbp]
  \caption{Named entity category+value pair detection results for E2E and  E2E* systems} 
  \label{tab:4G_val_etiquettes}
  \centering
  \begin{tabular}{ |l|c|c|c|c|}
    \hline
 \multicolumn{1}{|c|}{\textbf{System}} &\textbf{Corpus} & \multicolumn{1}{|c|}{\textbf{Precision}} & \multicolumn{1}{|c|}{\textbf{Recall}} & \multicolumn{1}{|c|}{\textbf{F-measure}}    \\
    \hline
    \hline
    E2E & dev   & \bf0.64  &  0.45& 0.53  \\
    E2E& test  & \bf0.55  & 0.36& 0.44  \\
    \hline
    \hline
    E2E* &dev   & 0.57 & \bf{0.47} &  0.52 \\
    E2E* &test  & 0.47&  \bf{0.38}& 0.42  \\
    \hline
  \end{tabular}
  
\end{table}




Last, we would like to compare these results to the ones obtained by a pipeline process, that consists in applying a text named entity recognition on the automatic transcripts produced by the end-to-end ASR system trained on the first step of the multi-task learning presented above.

The text named entity recognition system used for this experiment is based on the combination of bi-directional LSTM (BLSTM), CNN and CRF modules~\cite{ma2016end}, and takes benefits from both word and character-level embeddings learned automatically during the training process. For this experiment, we used the NeuroNLP2 implementation\footnote{https://github.com/XuezheMax/NeuroNLP2}.
Convolutional neural network encodes character-level information of a word into its character-level embedding. Then the character-and word-level embeddings are fed into the BLSTM to model context information of each word. On top of BLSTM, the sequential CRF is used to jointly decode labels for the whole sentence. 
In addition, this system can be enriched with syntactic information like part of speech tagging (POS).
In our experiment, NeuroNLP2 is used as a NER system and Deep Speech 2 as the ASR system. Both are trained on the DeepSUN corpus described in section~\ref{sec:corpus}.
Automatic transcriptions of dev and test data have been annotated with NER system. To measure the impact of POS, we used the MACAON system~\cite{nasr2011macaon} to tag of DeepSUN corpus and manual and automatic transcriptions. 

To feed NeuroNLP2, one hot vectors represent POS information. Word embeddings, character representations and one hot concatenations feed the BLSTM layer.
As we can see in Tables~\ref{tab:neuro_etiquettes_seules}, the pipeline process is less competitive than the end-to-end model to recognize EN category, but is more efficient to extract EN values. Results also confirms that linguistic information like POS is really important for the NER task. Such observation will help for future work on the continuity of this study.

\begin{table}[!htbp]
  \caption{NER results for the pipeline approach (Pip) on the \textbf{test} data. When POS are used to tag ASR outputs before NER processing, the system is called Pip+POS} 
  \label{tab:neuro_etiquettes_seules}
  \centering
  \begin{tabular}{| l|l|c|c|c| }
    \hline
    
    \multicolumn{1}{|c|}{\textbf{System}}  &\textbf{Detection} &\multicolumn{1}{|c|}{\textbf{Precision}} & \multicolumn{1}{|c|}{\textbf{Recall}} & \multicolumn{1}{|c|}{\textbf{F-measure}}    \\
    \hline
    \hline
    Pip & category  & \bf0.75 & 0.56 & 0.64  \\
    Pip+POS &category       & 0.74 & \bf0.58 & \bf0.65  \\
    \hline
    Pip &cat+value  & \bf0.58 & 0.43 & 0.49  \\
    Pip+POS &cat+value       & 0.57 & \bf0.45 & \bf0.50  \\
	\hline

  \end{tabular}
    \vspace{-0.5cm}
\end{table}



As described in section~\ref{sec:tricks}, we applied NeuroNLP2 (the version using POS tagging) on the manual transcriptions of the ASR training data in order to augment the amount of ''NER from speech'' training data. In this experiment,  the normal and starred modes  were  used. Table~\ref{tab:e2e_augment} shows the improvement got by the end-to-end system when training on these imperfect augmented data using the normal (E2E+) and the starred (E2E+*) modes.  As we can see, the use of the augmented data was helpful for the starred mode.

\begin{table}[!htbp]
  \caption{NER results on the \textbf{test} data for the E2E system trained with imperfect augmented data (E2E+) in comparison to the E2E system trained with imperfect augmented data and the starred mode (E2E+*)} 
  \label{tab:e2e_augment}
  \centering
  \begin{tabular}{ |l|l|c|c|c| }
    \hline
   \multicolumn{1}{|c|}{\textbf{System}} &\textbf{Detection} & \multicolumn{1}{|c|}{\textbf{Precision}} & \multicolumn{1}{|c|}{\textbf{Recall}} & \multicolumn{1}{|c|}{\textbf{F-measure}}    \\
  \hline  
  \hline
    E2E+ & category         & \bf 0.82 & \bf0.57 &   0.67  \\
    E2E+*& category  &  0.76&  0.63& \bf0.69  \\
    \hline
    E2E+ & cat+value       &\bf0.55 & 0.40 &  0.46  \\
    E2E+* & cat+value        & 0.49 & \bf0.41 & \textbf{0.47} \\
    \hline
  \end{tabular}
  \vspace{-0.5cm}
\end{table}

\section{Conclusion}
This paper presents a first study about end-to-end named entity extraction from speech. By integrating in the character sequence emitted by a CTC end-to-end speech recognition system some special characters to delimit and categorize named entities, we showed that such extraction is feasible. 
To compensate the lack of training data, we propose a multi-task learning approach (ASR + NER) in addition to an artificial data augmentation of the training corpus with automatic annotation of named entities. 
A starred mode is also proposed to make the neural network more focused on named entities. 
Experimental results show that this end-to-end approach in starred mode with training augmentation, provides better results (F-measure equals to 0.69 on test) than a pipeline approach to detect named entity categories (F-measure=0.64). On the other side, performances of this end-to-end approach to extract named entity values are worse than the ones got by the pipeline process. 

As a conclusion, this study presents promising results in a first attempt to experiment an end-to-end approach to extract named entities, and constitutes an interesting start point for future work that could start by combining starred mode with training data augmentation, but also explore more different ways, like injecting linguistic information in the end-to-end neural architecture.
\vspace{-0.2cm}
\section{Acknowledgements}
\small
This work was supported by the French ANR Agency through the CHIST-ERA M2CR project, under the contract number ANR-15-CHR2-0006-01, and by the RFI Atlanstic2020 RAPACE project.
Authors would like to sincerely thank Sean Naren to make his implementation of Deep Speech 2 available, as well as the contributors to the NeuroNLP2 project.
\bibliographystyle{IEEEtran}
\bibliography{mybib}
\end{document}